%% file: main.tex
\pdfoutput=1
\documentclass[english,letterpaper, 10 pt, conference]{iros}
\usepackage[T1]{fontenc}
\usepackage[latin9]{inputenc}
\usepackage{babel}
\usepackage{refstyle}
\usepackage{booktabs}
\usepackage{url}
\usepackage{amstext}
\usepackage{graphicx}
\usepackage{microtype}
\usepackage[unicode=true,pdfusetitle,
 bookmarks=true,bookmarksnumbered=false,bookmarksopen=false,
 breaklinks=false,pdfborder={0 0 1},backref=false,colorlinks=false]
 {hyperref}

\makeatletter

\AtBeginDocument{\providecommand\figref[1]{\ref{fig:#1}}}
\AtBeginDocument{\providecommand\subsecref[1]{\ref{subsec:#1}}}
\AtBeginDocument{\providecommand\eqref[1]{\ref{eq:#1}}}
\AtBeginDocument{\providecommand\secref[1]{\ref{sec:#1}}}
\AtBeginDocument{\providecommand\tabref[1]{\ref{tab:#1}}}
\providecommand{\tabularnewline}{\\}
\newcommand{\lyxdot}{.}

\RS@ifundefined{subsecref}
  {\newref{subsec}{name = \RSsectxt}}
  {}
\RS@ifundefined{thmref}
  {\def\RSthmtxt{theorem~}\newref{thm}{name = \RSthmtxt}}
  {}
\RS@ifundefined{lemref}
  {\def\RSlemtxt{lemma~}\newref{lem}{name = \RSlemtxt}}
  {}

\IEEEoverridecommandlockouts
\usepackage{tikz}

\usepackage[nolist]{acronym}
\newacro{BMS}[BMS]{Boolean Map Saliency}
\newacro{SOTA}[SOTA]{State-Of-The-Art}
\newacro{CSA}[CSA]{Center Surround Antagonism}
\newacro{ADAS}[ADAS]{Advanced Driver Assistance System}
\newacro{KP}[KP]{KeyPoint}

\usepackage[noadjust]{cite}
\usepackage{refstyle}

\newref{sec}{%
name = \RSsectxt,
names = \RSsecstxt,
Name = \RSSectxt,
Names = \RSSecstxt,
refcmd = {\ref{#1}},  
rngtxt = \RSrngtxt,
lsttwotxt = \RSlsttwotxt,
lsttxt = \RSlsttxt}

\newref{subsec}{%
name = \RSsectxt,
names = \RSsecstxt,
Name = \RSSectxt,
Names = \RSSecstxt,
refcmd = {\ref{#1}},  
rngtxt = \RSrngtxt,
lsttwotxt = \RSlsttwotxt,
lsttxt = \RSlsttxt}

\newref{appendix}{%
name = \RSappendixname,
names = \RSappendicesname,
Name = \RSAppendixname,
Names = \RSAppendicesname,
refcmd = {\ref{#1}},  
rngtxt = \RSrngtxt,
lsttwotxt = \RSlsttwotxt,
lsttxt = \RSlsttxt}

\usepackage{tikz}

\makeatother

\begin{document}
\title{\Large \bf Combining Visual Saliency Methods and Sparse Keypoint Annotations to Providently Detect Vehicles at Night}
\author{Lukas Ewecker,$^{\textrm{*, 1}}$\thanks{$^{\textrm{*}}$Authors contributed equally.}\thanks{$^{\textrm{1}}$Dr.\ Ing.\ h.c.\ F.\ Porsche AG, Weissach, Germany.
\texttt{\{lukas.ewecker, robin.schwager, stefan.roos4, tim.bruehl\}@porsche.de}} Lars Ohnemus,$^{\textrm{*, 2}}$\thanks{$^{\textrm{2}}$Karlsruhe Institute of Technology (KIT), Germany.
The research was performed during employment at Dr.\ Ing.\ h.c.\ F.\ Porsche\ AG\@.\texttt{
lars.ohnemus@student.kit.edu}} Robin Schwager,$^{\textrm{*, 1}}$ Stefan Roos,$^{\textrm{1}}$ Tim
Brühl,$^{\textrm{1}}$ and Sascha Saralajew$^{\textrm{*, 3}}$\thanks{$^{3}$NEC Laboratories Europe GmbH, Heidelberg, and Leibniz University
Hannover, Institute of Product Development, Hannover, both Germany.
The research was performed during employment at Dr.\ Ing.\ h.c.\ F.\ Porsche\ AG\@.
\texttt{sascha.saralajew@neclab.eu}}}

\maketitle
\setcounter{footnote}{3}

\input{00_abstract.tex}

\input{01_introduction.tex}

\input{02_related_work.tex}

\input{03_saliency_generation.tex}

\input{05_experiments.tex}

\input{06_discussion_conclusion.tex}


\bibliographystyle{IEEEtran}
\bibliography{paper}

\end{document}

%% file: 00_abstract.tex
\begin{abstract}
Provident detection of other road users at night has the potential
for increasing road safety. For this purpose, humans intuitively use
visual cues, such as light cones and light reflections emitted by
other road users to be able to react to oncoming traffic at an early
stage. Computer vision methods can imitate this behavior by predicting
the appearance of vehicles based on emitted light reflections caused
by the vehicle's headlights. Since current object detection algorithms
are mainly based on detecting directly visible objects annotated via
bounding boxes, the detection and annotation of light reflections
without sharp boundaries is challenging. For this reason, the extensive
open-source PVDN (Provident Vehicle Detection at Night) dataset was
published that includes traffic scenarios at night with light reflections
annotated via keypoints. In this paper, we explore a new approach
to annotate objects without clear boundaries, such as light reflections,
by combining sparse keypoint annotations of humans with the concept
of Boolean map saliency. With that, we create context-aware saliency
maps that capture the inherently unsharp boundaries of light reflections.
We show that this approach allows for an automated derivation of different
object representations, such as saliency maps or bounding boxes, so
that detection models can be trained on different annotation variants
and the problem of providently detecting vehicles at night can be
tackled from different perspectives. Our approach makes it possible
to derive bounding boxes with superior quality compared to previous
approaches and develop better object detection algorithms. With this
paper, we provide additional powerful tools and methods to study the
problem of detecting vehicles at night before they are actually visible.
\end{abstract}

%% file: 01_introduction.tex
\section{Introduction\label{sec:Introduction}}


Current State-Of-The-Art (SOTA) methods in object detection mostly
rely on annotations via bounding boxes. This is a valid approach as
objects commonly considered in such tasks are easily describable by
bounding boxes. A car in daylight, for example, has clear and objectively
definable object borders. This makes it very easy for human annotators
to draw a bounding box around a car, as they have no problem identifying
the car's boundaries. However, objectively drawing bounding boxes
becomes a challenge when dealing with objects that are not described
by clear, sharp boundaries. For instance, when an object seemingly
fades into the background, it is difficult to set rules for manual
annotation, which ultimately leads to a high degree of uncertainty
in the annotations of different annotators and can result in highly
variable annotations. In particular, this phenomenon becomes evident
when dealing with the problem of provident vehicle detection at night.
The problem was first brought up by Oldenziel et al.~\cite{Oldenziel2020}
and recently Saralajew et al.~\cite{saralajew2021pvdn-dataset} published
the corresponding large-scale PVDN (Provident Vehicle Detection at
Night) dataset, where oncoming vehicles and their corresponding light
reflections are annotated. Here, the objective labeling of light reflections
with a fuzzy nature and soft borders emerged as a problem. To this
end, they conducted a study comparing the consistency of different
annotators via bounding boxes for the same images. They conclude that
the expert annotators could not agree on a consistent ground truth,
which can cause problems for systems learning from this data. On the
other side, the importance of clean and consistent ground truth for
deep learning tasks has been demonstrated by several researchers~\cite{chadwick2019noisydata,frenay2014noisydata,ma2022noisydata,schilling2022noisydata}.
Consequently, the dataset was not annotated using bounding boxes but
instead using keypoints placed on the point with the highest intensity
of each light artifact. This allows for a much more objective annotation
and leaves room to automatically derive rule-based object representations,
such as bounding boxes. Saralajew et al.~\cite{saralajew2021pvdn-dataset}
already mentioned the possibility of using saliency-based approaches
in order to objectively derive saliency maps from the keypoint annotations
and to use them to generate different object representations, such
as binary maps or bounding boxes. However, without further exploring
the saliency-based approaches in detail.
\begin{figure*}[t]
\begin{centering}
\includegraphics[width=0.8\paperwidth]{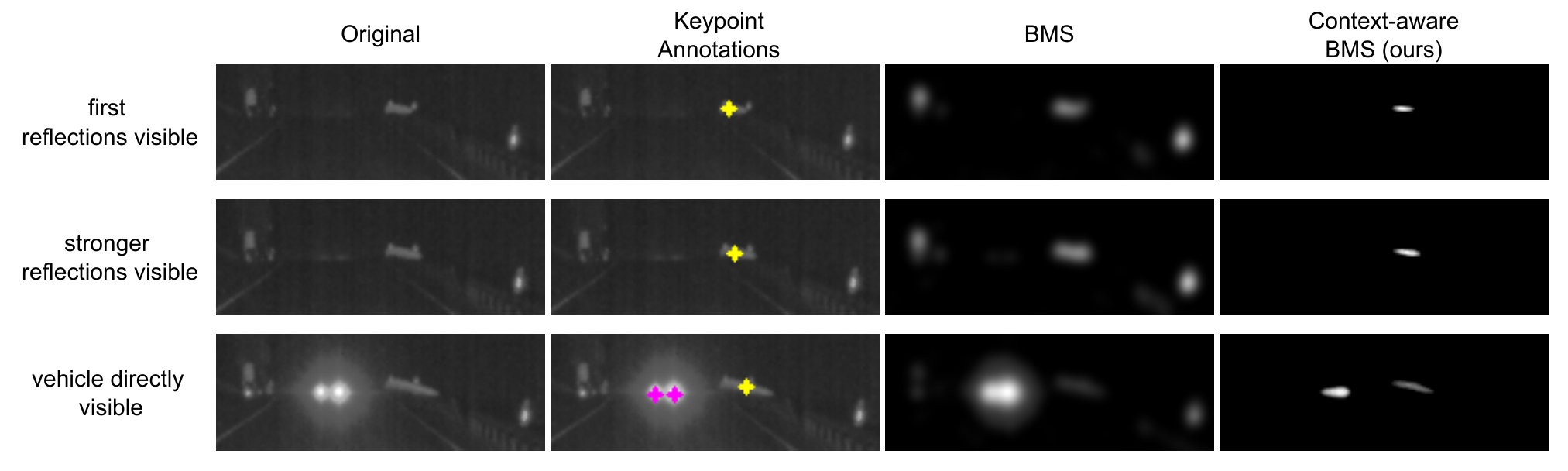}
\par\end{centering}
\caption{\label{fig:bms-comparison}Visualization of the difference between
the original BMS method~\cite{zhang2013} and our proposed context-aware
BMS variant. The three images in the first column show a scene from
the PVDN dataset~\cite{saralajew2021pvdn-dataset} from the point
where the first light reflections are visible---in this case on the
guardrail---until the vehicle becomes directly visible itself. The
second column shows the annotated ground-truth keypoints for each
light instance, where the pink and yellow represent direct and indirect
light instances, respectively. The third and fourth columns show the
saliency maps with the original BMS and our context-aware BMS method,
respectively. The example clearly shows that the original BMS method
still outputs other irrelevant light instances, such as illuminated
street signs, as salient regions, whereas our approach only marks
the relevant light instances and accurately captures their soft borders.}
\end{figure*}

In this paper, we study an approach to create an objective representation
of fuzzy objects (such as light reflections) that accurately captures
unsharp and soft boundaries given keypoint annotations. For this purpose,
we extend the attention map concept of the Boolean Map Saliency (BMS)
method~\cite{zhang2013} by combining it with sparse keypoint annotations
by humans. With that, we are able to create context-aware saliency
maps that annotate relevant light artifacts in images without requiring
clear object boundaries. An illustration of the results using our
approach compared to the original BMS method is given in \figref{bms-comparison}.
Using the proposed approach, we show how to automatically derive various
object representations such as saliency maps and bounding boxes. Moreover,
we analyze the advantages and limitations of this method and show
it is a suitable approach to annotate objects with fuzzy borders objectively.
Finally, we study the suitability of our approach as a method to automatically
generate bounding boxes in order to train SOTA object detection algorithms.
With these contributions, we want to provide further tools and methods
to work with the PVDN dataset and want to enable fellow researchers
to study the problem of providently detecting vehicles at night. To
this end, we publish the source code of the proposed method.\footnote{\url{https://github.com/lukazso/kpbms-pvdn/}}
Beyond the studied application, we are confident that the proposed
method can be applied to all areas where one wants to detect or annotate
objects with fuzzy object boundaries (as in medical images).

The paper is organized as follows: First, we provide a brief overview
of the related work on the topics of visual saliency and provident
vehicle detection at night. Second, we explain the proposed saliency
method. After that, we present experiments where we applied the method
for the task of provident vehicle detection at night. We finish the
paper with a conclusion and outlook.

%% file: 02_related_work.tex
\section{Related Work\label{sec:Related-Work}}

\subsection{Visual Saliency}

Despite its long history, visual saliency and salient object detection
are active fields of research~\cite{Ullah2020}. There are many different
approaches to mimicking the human attention mechanism, both heuristic
and learning-based. For example, Itti et al.~\cite{Itti1998,Itti1999}
presented a human attention model that uses three different feature
maps---color, orientation, and intensity---and center-surround mechanisms
at different scales to generate saliency maps. Newer research focuses
both on conventional computer vision~\cite{zhang2013,Jian2018} as
well as deep learning approaches~\cite{Zhang2017,pan2016shallow,Kroner2020,liu2021saliency-transformer}
to generate saliency maps and detect salient objects. Saliency maps
can be used for various applications in the field of computer vision~\cite{Ullah2020}.
Most prominently, saliency maps can be used to infer segmentation
masks for salient object segmentation as well as classify salient
objects~\cite{Li_2017_CVPR}. Other use cases are visual tracking
and video compression~\cite{Ullah2020}. 

The performance of such saliency models can be evaluated using eye
tracking data: A sensor tracks eye fixation points within an image~\cite{borji2015cat2000}.
Such fixture points are related to keypoint annotations with the difference
that keypoints have clear associated features (e.\,g., human joints)~\cite{Andriluka2014,belagiannis2017recurrent,carreira2016human}.

Visual saliency can also be applied in the automotive context to mimic
the human perception apparatus~\cite{Maldonado2021,Lateef2021}.
Pugeault and Bowden~\cite{Pugeault2015} investigated how much of
human behavior during driving happens pre-attentive---and, therefore,
at a more fundamental level in the human visual perception. They also
showed that current bottom-up saliency methods are not expressive
enough to be successfully used in the automotive context. However,
this discovered human superiority suggests potential for elaborating
better models of human perception, especially at night, when this
discrepancy is even more visible since visual perception is more difficult. 

\subsection{Provident Vehicle Detection at Night}

At night, the human provident, pre-attentive behavior is still unchallenged
by advanced driver assistance systems, which are currently mostly
relying on visible headlights of other cars to detect them~\cite{Lopez.2008,P.F.Alcantarilla.2011,Eum.2013,Juric.2014,Sevekar.2016,Satzoda2019}.
Therefore, early signs of oncoming vehicles, like local light reflections
on guardrails and the street, are unused features that humans highly
rely on. Oldenziel et al.~\cite{Oldenziel2020} studied the deficit
between human provident vehicle detection and a camera-based vehicle
detection system. Based on a test group study, they showed that drivers
detected oncoming vehicles on average 1.3\,s before the vehicle is
actually directly visible, which is a not negligible time discrepancy.
Ewecker et al.~\cite{ewecker2022iv} showed that for urban scenarios,
the average time human drivers detect oncoming vehicles before direct
sight is 0.3\,s. Tests showed that deep learning predictors can detect
oncoming vehicles based on light artifacts to some extent~\cite{Oldenziel2020,saralajew2021pvdn-dataset}.
But, in the analysis of Saralajew et al.~\cite{saralajew2021pvdn-dataset},
they raised concerns about the ability to describe fuzzy light artifacts
through rigid bounding boxes used in the evaluated predictors. As
a consequence, they published the PVDN dataset\footnote{\url{https://doi.org/10.34740/KAGGLE/DS/1061422}}
containing around 60\,000 annotated gray-scale images~\cite{saralajew2021pvdn-dataset}.
The dataset captures rural environments with a commonly used automotive
front camera. There, all \emph{light instances} (light artifacts)
caused by oncoming vehicles are annotated. Those light instances are
categorized into \emph{direct} (e.\,g., headlights) and \emph{indirect
light instances} (e.\,g., light reflections on guardrails caused
by the oncoming vehicle). In contrast to Oldenziel et al.~\cite{Oldenziel2020},
the PVDN dataset uses keypoint annotations that capture human attention
through a clear annotation hierarchy to allow the investigation of
several use cases, for example, regressing more objective bounding
box representations. Moreover, with the keypoints as initial seeds,
the authors further explored methods to extend the keypoint annotations
to bounding boxes with low annotation uncertainty and successfully
trained both shallow and deep learners on the regressed annotations.
The bounding box regression in their work is based on a simple thresholding
schema. Here, the entire image is firstly binarized using an adaptive
thresholding technique. Then, bounding boxes are inferred from the
thresholded image. Lastly, bounding boxes containing no keypoint annotations
are discarded as false positives. They also addressed the possibility
of combining their sparse keypoint annotations with visual saliency
methods to enhance the information content of the PVDN data. We pick
this idea up and provide a simple tool to extend the base keypoint
annotations. For simplicity, we will use the terms ``keypoints''
and ``light instances'' interchangeably. 

\subsection{Annotation of Unsharp Objects}

Conventional object detection tasks---especially in autonomous driving
or robotic applications---normally do not suffer from the problem
of unsharp and soft borders. Consequently, current SOTA object detection
datasets contain objects in the classical sense with objectively definable
boundaries~\cite{caesar2020nuscenes,geiger2012kitti,coco,yu2020bdd100k}.
For example, given a conventional object such as a car in daylight,
for humans, it can be considered a fairly easy task to separate the
car from its background based on its clear object boundaries. However,
this becomes difficult when annotating unconventional objects, such
as light reflections. There, relevant light artifacts change their
shape arbitrarily based on the reflection surface and lack clearly
definable boundaries as reflections often vanish seamlessly into the
background. Saralajew et al.~\cite{saralajew2021pvdn-dataset} showed
the ambiguity of human bounding box annotations in terms of object
size when asked to annotate light reflections caused by oncoming vehicles
at night. In this case, keypoints placed on the point of a reflection's
highest intensity were chosen instead of bounding boxes as the final
annotation methodology. With that, the authors avoid the challenge
of describing the exact shape and size of a light instance. Besides
automotive applications, medical images also often cover objects with
fuzzy shapes~\cite{codella2018skin,lee2017mamography-dataset,Smal2010,Wu2015,Abousamra2018}.
For instance, lesions, tumors, or various objects in fluorescent microscopy
can often appear as unsharp blobs. However, current annotation methods
in those applications are mostly limited to binary masks or bounding
boxes. Objective annotations are thereby approached by optimizing
the annotation process via crowd annotations by non-experts~\cite{mckenna2012medical-annotation,nguyen2012medical-annotation},
expert annotations~\cite{lee2017mamography-dataset}, or a combination
of both~\cite{albarqouni2016medical-annotation}. However, no work
exists that focuses on annotation methods capturing objects with soft
borders like light artifacts. With our proposed method, we make a
first contribution to properly model the inherent characteristics
of unsharp objects. Although we study our proposed method based on
an automotive use case, we believe that it can be transferred to and
be of use for other domains---such as medical applications---as
well.

%% file: 03_saliency_generation.tex
\section{Attention Generation Method\label{sec:Saliency-Generation-Method}}

\begin{figure*}
\begin{centering}
\input{method_figure/method_figure.tex}
\par\end{centering}
\caption{\label{fig:Comparison-of-the}Comparison of the two workflows: The
upper part shows the workflow of the original BMS method~\cite{zhang2015exploiting}
for reference, while the lower part shows the adapted workflow (our
method). The example image used is from the PVDN dataset (ID:~017358,~\cite{saralajew2021pvdn-dataset}).}
\end{figure*}
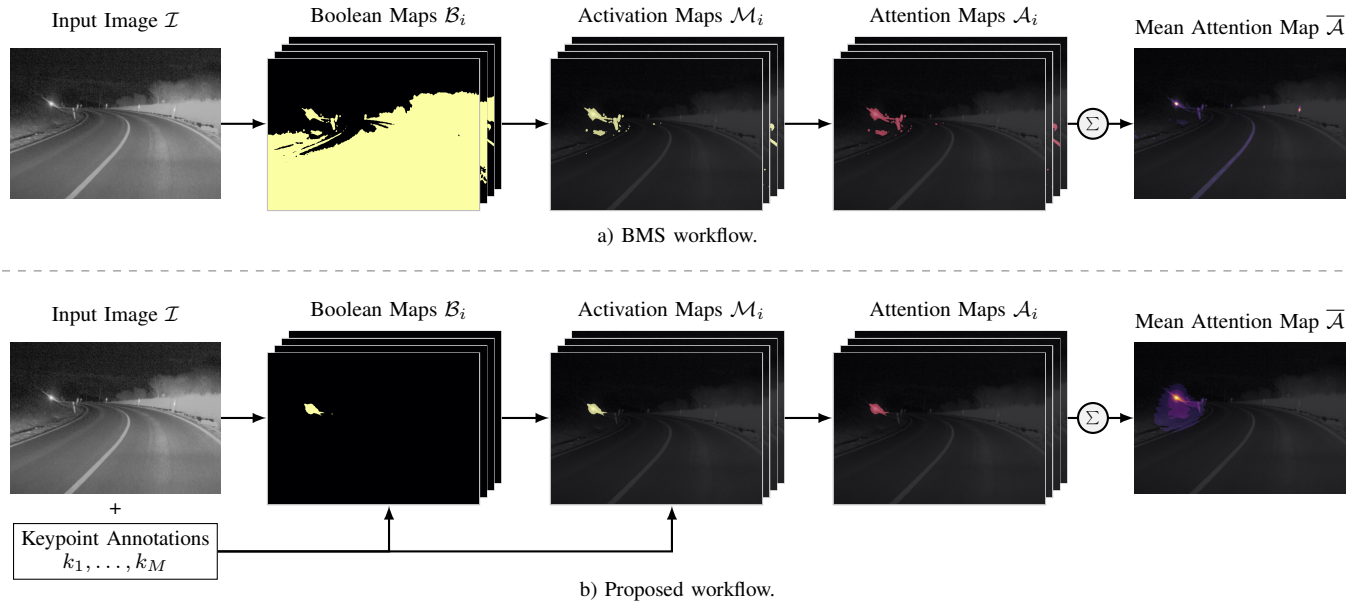
Visual saliency offers a mechanism to extract pertinent regions of
interest from images. This attention mechanism is purely based on
local topography and is not biased towards a specific problem. However,
when combined with sparse human annotations, the salient regions can
be filtered to assert context upon the resulting saliency maps. Then,
these maps provide more information than both methods alone could.
This emergence is exploited by the context-aware attention generation
method that we propose. In the following, we introduce the method
and specify an algorithm to generate context-aware saliency maps based
on the BMS method. 

\subsection{Saliency Method}

Saliency detection is a well-studied task~\cite{Zhang2017,pan2016shallow,Kroner2020,liu2021saliency-transformer,zhang2013,Jian2018}.
A simple and effective method to detect saliency is to use human perception
mechanisms, for instance, figure-ground segregation. One factor likely
influencing this segregation is the surroundedness of salient objects
by background~\cite{zhang2015exploiting}. Zhang and Sclaroff~\cite{zhang2013}
formulated the BMS method around  this assumed relationship. Their
method binarizes a given image at different levels with randomly sampled
thresholding values, generating so-called Boolean maps. These Boolean
maps are hypothesized to characterize the image and its salient regions~\cite{zhang2015exploiting}.
To distinguish salient patches, surrounded contours are activated
by masking regions in the Boolean map that are connected to the image's
borders~\cite{zhang2015exploiting}. The attention and thus saliency
is then computed by averaging the activated Boolean maps~\cite{zhang2015exploiting}.
Due to the simplistic foreground definition through the binary images,
this method is well suited for measuring connectedness between foreground
(saliency method) and the context-aware keypoint annotations. Also,
the saliency maps can be computed without further preliminary steps
(e.\,g., training), which simplifies the generation procedure. Therefore,
we will use the BMS method as a foundation for our compound approach.\footnote{Note that other saliency methods are suitable as well.}
In particular, our propose method is adapted from BMS by using a different
formulation of the Boolean and attention map generation (\subsecref[s]{Boolean-Map-Generation}
and~\ref{subsec:Attention-Map-Computation}). Instead of only using
the raw image data as information, a sparse set of keypoints is provided
through human annotations. These points are located in regions that
a human annotator judged interesting for a specific problem---for
instance, such an annotation technique was used by Saralajew et al.~\cite{saralajew2021pvdn-dataset}
in the PVDN dataset. Using the keypoints, the attention mechanism
can be rephrased: If a patch in a Boolean map contains a keypoint,
it is considered to be of interest and, therefore, salient (see \subsecref{Attention-Map-Computation}
for further details). This mechanism is incorporated in the saliency
generation in the following.

In the following paragraphs, a similar nomenclature and workflow to
Zhang and Sclaroff~\cite{zhang2015exploiting} is presented for easy
comparison. \figref{Comparison-of-the} shows both methods side-by-side.

\subsection{Image and Keypoints}

The input to the algorithm is assumed to be a grayscale image $\mathcal{I}\in\left[0,1\right]{}^{h\times w}$
of height $h$ and width $w$ and a set $K=\left\{ k_{j}=\left(x_{j},y_{j}\right):j=1,\ldots,M\right\} $
of $M$ sparse keypoints $k_{j}$ with $x_{j}\in\left\{ 1,\dots,w\right\} $
and $y_{j}\in\left\{ 1,\dots,h\right\} $. The keypoints are considered
to be annotated at highly salient spots in the image $\mathcal{I}$.
Importantly, the number $M$ of keypoints is small and, therefore,
common saliency inference from fixture points cannot be applied due
to the sparse distribution. Also, we expect an inherent relationship
between the intensity values $\phi_{j}:=\mathcal{I}_{y_{j},x_{j}}$
of the image $\mathcal{I}$ at the position of the keypoint $k_{j}=\left(x_{j},y_{j}\right)$
and the overall dynamics of the salient patch at this position. Following
the BMS method~\cite{zhang2015exploiting}, we assume that salient
patches have high intensities and that the intensity value $\phi_{j}$
is an approximation of a local intensity maximum.

The original BMS method is proposed for multi-channel images like
RGB\@. However, since the PVDN dataset consists of grayscale images,
we focus in the following explanation on images with one feature channel.
Nonetheless, like the BMS method, the proposed method is also extendable
to multi-channel images by repeating the steps ``Boolean map generation''
and ``attention map generation'' for every feature channel of an
image and by averaging over all channels afterwards or by applying
a respective channel transformation.

\subsection{Boolean Map Generation\label{subsec:Boolean-Map-Generation}}

The generation of a Boolean map $\mathcal{B}_{i}$, whereby $i=1,\ldots,N$
with $N$ being the number of sampled thresholds, is performed analogously
to Zhang and Sclaroff~\cite{zhang2013} by binarizing the input image
$\mathcal{I}$ at a randomly sampled threshold $\theta_{i}$:
\begin{equation}
\mathcal{B}_{i}=\mathtt{THRESH}\left(\mathcal{I},\theta_{i}\right).
\end{equation}
Using the intensity value $\phi_{j}$ at the position of the keypoint,
the number of sampled thresholds can be reduced to speed up the computation
process. Per definition, thresholds higher than the intensity value
at the keypoint position have no significant effect on the region's
saliency. For the same reason, significantly lower thresholds can
be ignored too. Consequently, the threshold $\theta_{i}$ can be sampled
uniformly from the interval $\left[\alpha\cdot\phi_{j},\phi_{j}\right]$,
where $\alpha\in[0,1]$ is a hyperparameter of the generation method.
The differences in the Boolean map generation process between our
context-aware approach and the traditional BMS are shown in \figref{Comparison-of-the}.

\subsection{Attention Map Computation\label{subsec:Attention-Map-Computation}}

The attention mechanism varies from the original BMS algorithm. Instead
of defining surroundedness by masking regions connected to the borders
in a Boolean map $\mathcal{B}_{i}$, we compute the activation map
$\mathcal{M}_{i}$ by using the set of keypoints $K$ to add context
information. Importantly, as already said, for the studied use case,
we assume keypoint annotations that are always placed in regions of
high interest and, therefore, at positions with high intensity values.
If a region in a Boolean map is connected to one of the keypoints,
it should therefore also be of interest. This fact is modeled by activating
the Boolean map with the keypoints as seeds instead of the image's
borders. This process can be formulated by
\begin{equation}
\mathcal{M}_{i}=\bigcup_{j=1}^{M}\mathtt{FLOOD}\left(\mathcal{B}_{i},k_{j}\right).\label{eq:activation_kpbms}
\end{equation}
The function $\mathtt{FLOOD}\left(\mathcal{B}_{i},k_{j}\right)$ applies
a flood-fill algorithm to $\mathcal{B}_{i}$ with $k_{j}$ as seed
position and, therefore, returns a map containing only regions connected
to the seed. As a consequence, less patches are activated (see \figref{Comparison-of-the}).
Note that this formulation does not rely on the position of the salient
regions within the image because activated regions can be in contact
with the image border. This is not true for the original BMS method
since it relies on the assumption that the image borders themselves
do not contain salient regions so that salient regions must be centrally
placed---the so-called Center Surround Antagonism (CSA)~\cite{zhang2015exploiting}.
However, in many applications, this CSA assumption cannot be adopted,
and rather the contrary case is of interest. For example, when detecting
unknown objects, their location can be everywhere in the image, and
assuming a prior centered distribution reduces the expressiveness
of the computed visual saliency. If the CSA can be assumed, an additional
computation step can be performed to enhance the resulting activation
maps by intersecting the activation map with the calculated BMS activation
map $\mathcal{M}_{i,\text{BMS}}$:
\begin{equation}
\mathcal{M}_{i,\text{comb}}=\mathcal{M}_{i,\text{BMS}}\cap\mathcal{M}_{i}.
\end{equation}
For the following examples, we will only use the first definition
from \eqref{activation_kpbms} because the CSA cannot be assumed for
the PVDN dataset~\cite{saralajew2021pvdn-dataset}. Also, note that
in contrast to the original BMS method, we do not split the activation
maps into sub-activation maps $\mathcal{M}_{i}^{+}$ and $\mathcal{M}_{i}^{-}$.
This has two reasons: First, for thresholds below the keypoint intensity,
the keypoint will always have a value of 1.0 in the corresponding
Boolean map $\mathcal{B}_{i}$ and, thus, $\mathcal{M}_{i}^{-}$ will
mostly be empty. Second, for higher intensities, the keypoint is likely
connected to large image regions, which causes ``leakage'' of the
computed attention. Also, no dilation operation is performed on the
activation map $\mathcal{M}_{i}$ since we found no real benefit for
the studied application. Consequently, the attention map $\mathcal{A}_{i}$
is simply calculated by 
\begin{equation}
\mathcal{A}_{i}=\frac{\mathcal{M}_{i}}{\left\Vert \mathcal{M}_{i}\right\Vert _{2}}.
\end{equation}
As a final step, all attention maps are averaged into a mean attention
map $\overline{\mathcal{A}}$ by 
\begin{equation}
\overline{\mathcal{A}}=\frac{1}{N}\sum_{i=1}^{N}\mathcal{A}_{i},
\end{equation}
as described by Zhang and Sclaroff~\cite{zhang2015exploiting}. The
mean attention map then represents the final saliency map. If multiple
keypoint classes exist, the steps above can be performed for each
class separately, yielding one mean attention map per class. For example,
this is used for the generation of attention maps for direct and indirect
keypoint annotations of the PVDN dataset, as presented in the following.

%% file: method_figure/method_figure.tex
\usetikzlibrary{positioning,shapes.multipart, fit,backgrounds,calc}

\pgfdeclarelayer{bg}    
\pgfsetlayers{bg,main}  

\def\width{2.8cm}
\def\sepIm{0.5cm}

\tikzset{>=latex}

\newcommand{\stackedImages}[1]
{%
	\begin{tikzpicture}
		\foreach \X [count=\Z]in {04,03,02,01}
			{\node[rectangle, draw=lightgray, inner sep=0pt, line width=1pt] at (0,0,\Z/4) {\includegraphics[width=\width, viewport=120bp 150bp 750bp 600bp, clip]{#1_\X.png}};}
	\end{tikzpicture}
}

\tikzset{every label/.style={black, font=\fontsize{8}{8.4}\selectfont, align=center, text width=\width}}

\begin{tikzpicture}[
	marrow/.style = {->, black, thick},
	imageNode/.style = {anchor=west, inner sep=0pt}
]

	\node[inner sep=1mm, label= Input Image $\mathcal{I}$](input_bms) at (0,0)
	{%
		\includegraphics[width=\width, viewport=120bp 150bp 750bp 600bp, clip]{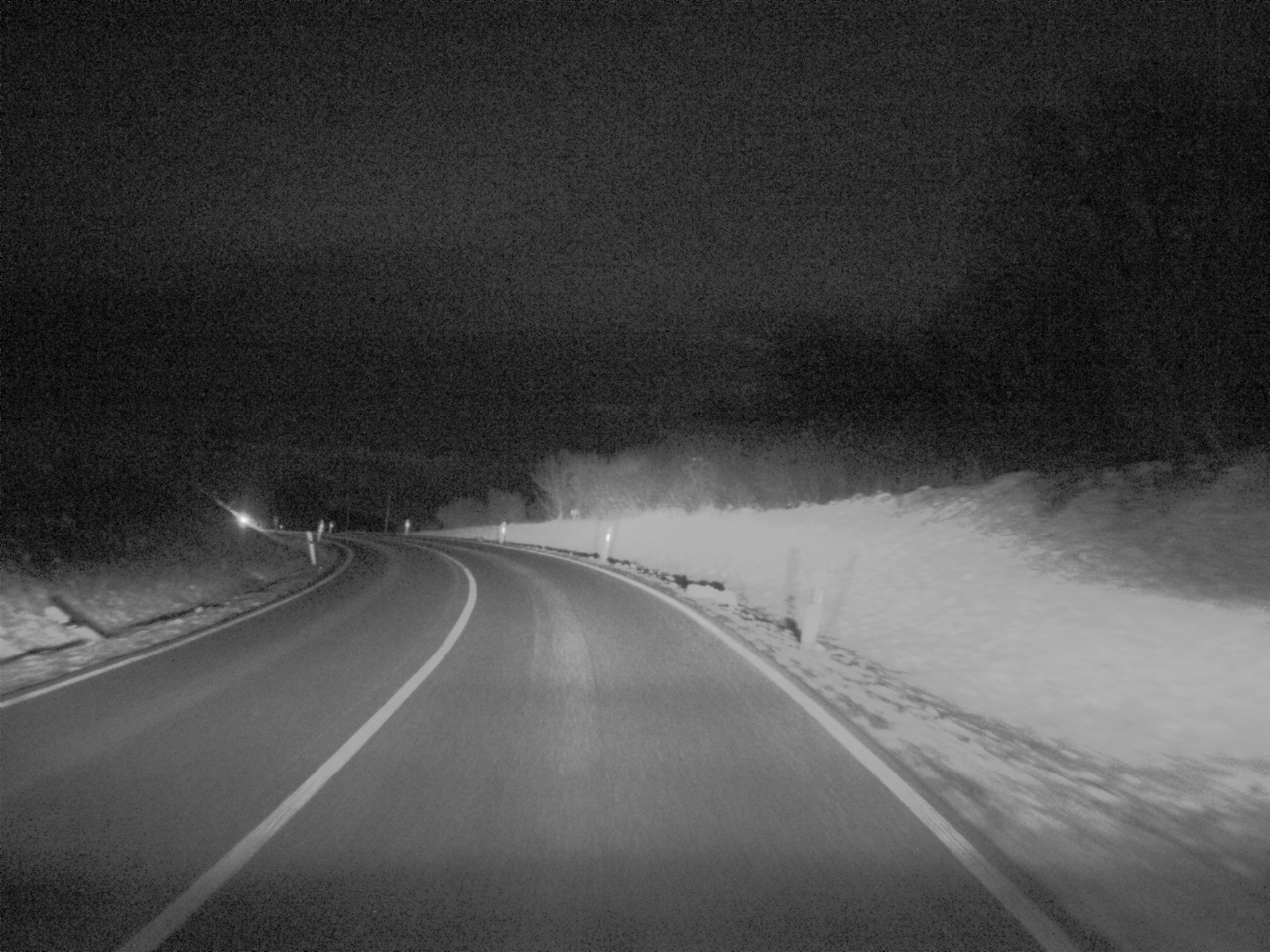}
	};
	\node[imageNode, label=Boolean Maps $\mathcal{B}_i$](bmaps_bms) at ($(input_bms.east) + (\sepIm, 0)$)
	{%
		\stackedImages{method_figure/bms/bool_map}
	};
	\node[imageNode, label=Activation Maps $\mathcal{M}_i$](mmaps_bms) at ($(bmaps_bms.east) + (\sepIm, 0)$)
	{%
		\stackedImages{method_figure/bms/act_map}
	};
	\node[imageNode, label=Attention Maps $\mathcal{A}_i$](amaps_bms) at ($(mmaps_bms.east) + (\sepIm, 0)$)
	{%
		\stackedImages{method_figure/bms/att_map}
	};
	\node[imageNode, label= Mean Attention Map $\overline{\mathcal{A}}$](out_bms)  at ($(amaps_bms.east) + (1.5*\sepIm, 0)$)
	{%
		\includegraphics[width=\width, viewport=120bp 150bp 750bp 600bp, clip]{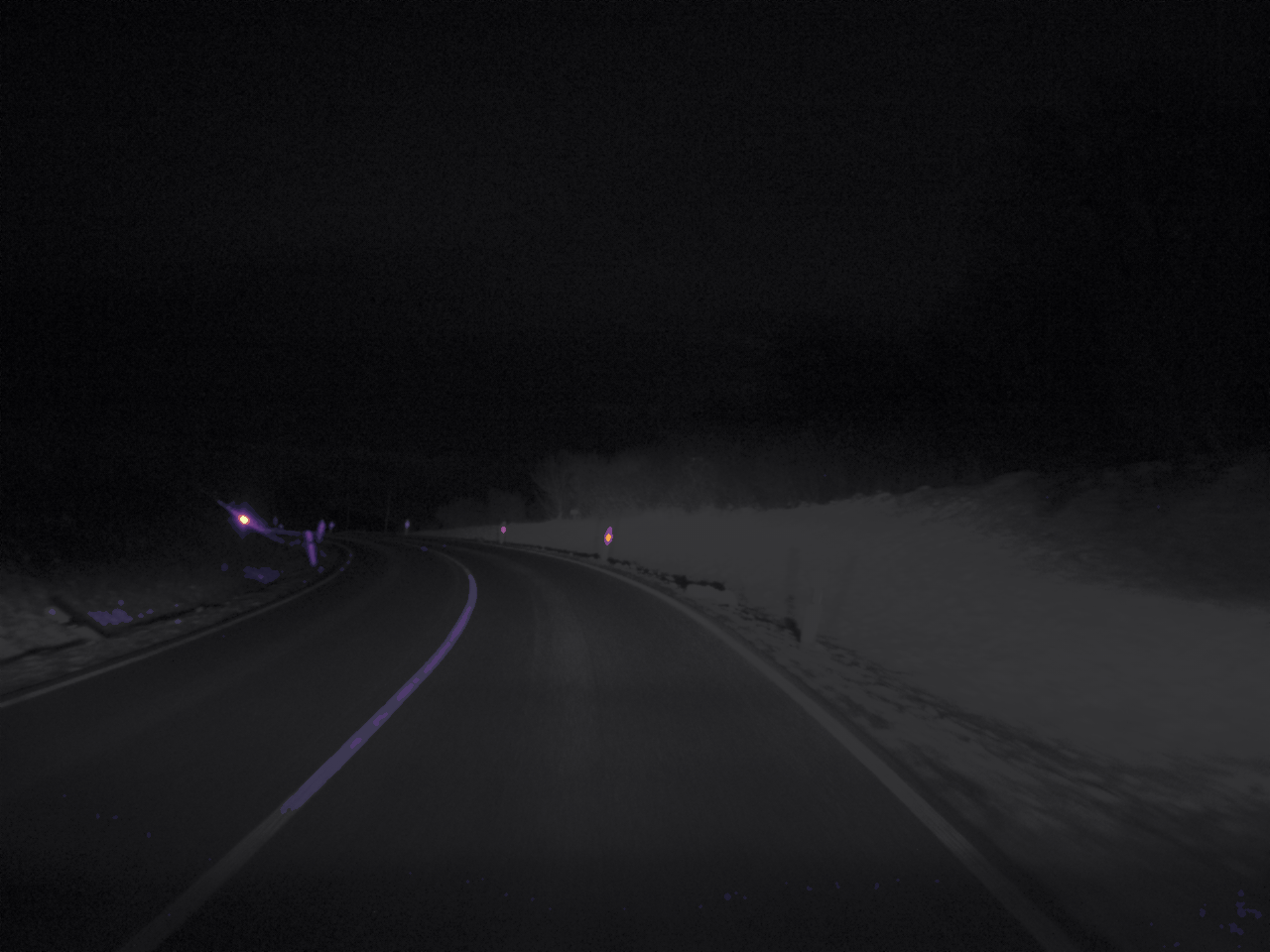}
	};

	\node[anchor=north, inner sep=1mm, label= Input Image $\mathcal{I}$](input_kpbms) at ($(input_bms.south) + (0, -1.7cm)$)
	{%
		\includegraphics[width=\width, viewport=120bp 150bp 750bp 600bp, clip]{method_figure/input_image.png}
		
	};
	\node[anchor=north, font=\fontsize{8}{8.4}\selectfont](plus_kp) at ($(input_kpbms.south) + (0, 1mm)$){+};
	\node[rectangle, draw, anchor=north, inner sep=1mm, font=\fontsize{8}{8.4}\selectfont, align=center](kp_kpbms) at ($(plus_kp.south) + (0, 0)$)
	{%
		Keypoint Annotations \\ 
		$k_1,\ldots,k_M$
	};

	\node[imageNode, label=Boolean Maps $\mathcal{B}_i$](bmaps_kpbms) at ($(input_kpbms.east) + (\sepIm, 0)$)
	{%
		\stackedImages{method_figure/kpbms/bool_map}
	};
	\node[imageNode, label=Activation Maps $\mathcal{M}_i$](mmaps_kpbms) at ($(bmaps_kpbms.east) + (\sepIm, 0)$)
	{%
		\stackedImages{method_figure/kpbms/act_map}
	};
	\node[imageNode, label=Attention Maps $\mathcal{A}_i$](amaps_kpbms) at ($(mmaps_kpbms.east) + (\sepIm, 0)$)
	{%
		\stackedImages{method_figure/kpbms/att_map}
	};
	\node[imageNode, label=Mean Attention Map $\overline{\mathcal{A}}$](out_kpbms)  at ($(amaps_kpbms.east) + (1.5*\sepIm, 0)$)
	{%
		\includegraphics[width=\width, viewport=120bp 150bp 750bp 600bp, clip]{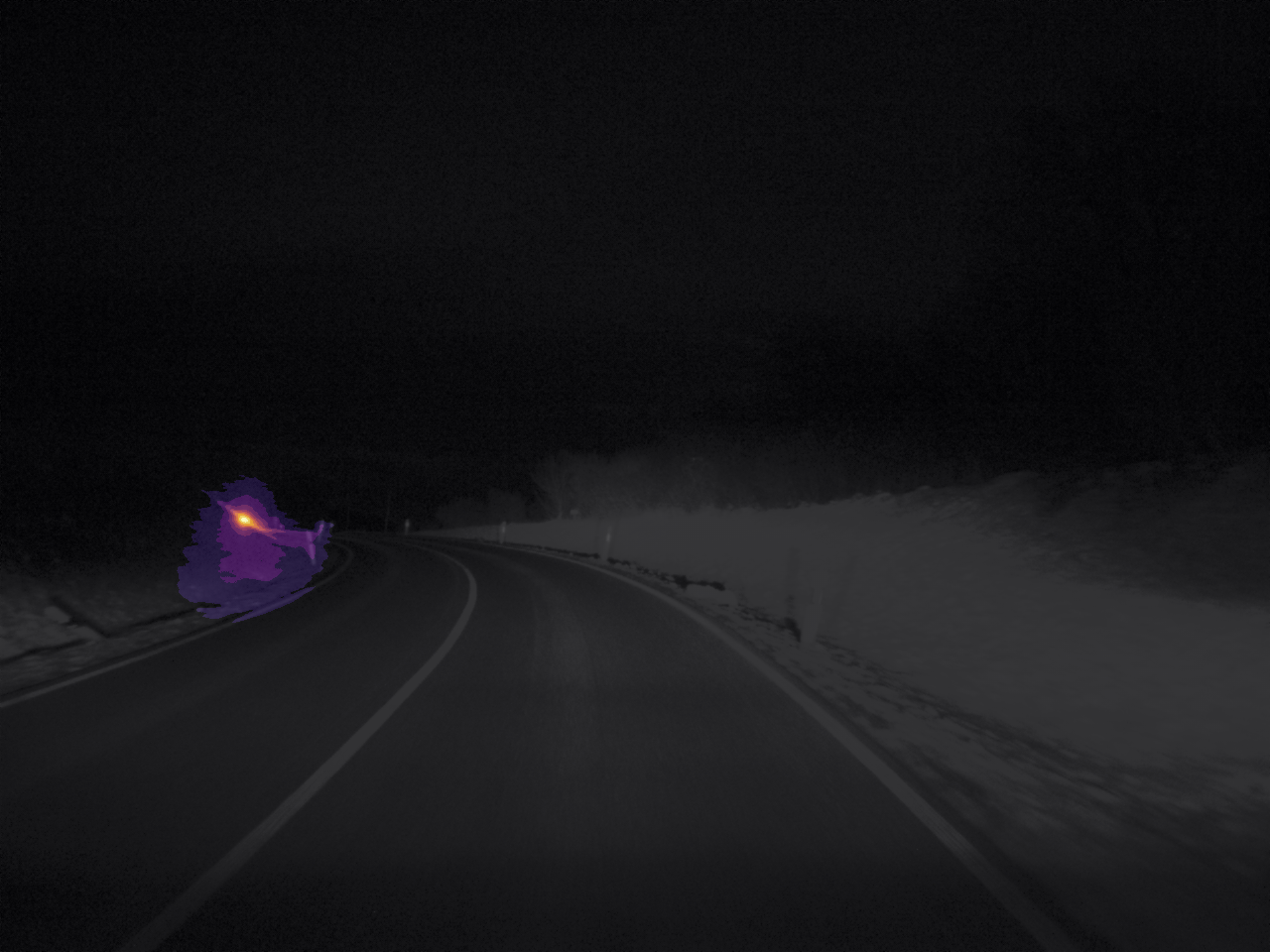}
	};

	\coordinate (center1) at ($(input_bms.south)!0.5!(out_bms.south)$);
	\coordinate (center2) at (kp_kpbms.south -| center1);
	\node[yshift=-2mm, anchor=north, font=\fontsize{8}{8.4}\selectfont] at (center1){a) BMS workflow.};
	\node[yshift= 1mm, anchor=north, font=\fontsize{8}{8.4}\selectfont] at (center2){b) Proposed workflow.};

	\begin{pgfonlayer}{bg}
		\draw[marrow] (input_bms.center) 	-- (bmaps_bms);
		\draw[marrow] (bmaps_bms.center) 	-- (mmaps_bms);
		\draw[marrow] (mmaps_bms.center) 	-- (amaps_bms);
		\draw[marrow] (amaps_bms.center) 	-- (out_bms);
		\draw[marrow] (kp_kpbms.east)		-| (bmaps_kpbms.south);
		\draw[marrow] (kp_kpbms.east)		-| (mmaps_kpbms.south);
		\draw[marrow] (input_kpbms.center) 	-- (bmaps_kpbms);
		\draw[marrow] (bmaps_kpbms.center) 	-- (mmaps_kpbms);
		\draw[marrow] (mmaps_kpbms.center) 	-- (amaps_kpbms);
		\draw[marrow] (amaps_kpbms.center) 	-- (out_kpbms);
	\end{pgfonlayer}

	\coordinate[] (p1) at ($(input_bms.north west)!.5!(input_kpbms.south west)$);
	\coordinate (p2) at (p1 -| out_bms.south east);
	\draw[dashed, draw=gray] (p1) -- (p2);

	\draw node[anchor=west, circle, draw, thick, fill=lightgray!20, inner sep=1mm, scale=0.5](sum_bms) at ($(amaps_bms.east) + (0, 0)$)
	{%
		$\sum$
	};
	\draw node[anchor=west, circle, draw, thick, fill=lightgray!20, inner sep=1mm, scale=0.5](sum_bms) at ($(amaps_kpbms.east) + (0, 0)$)
	{%
		$\sum$
	};
	
\end{tikzpicture}

%% file: 05_experiments.tex
\section{Experiments\label{sec:Experiments}}

To evaluate the proposed context-aware BMS method, we conduct experiments
on the PVDN dataset. First, we show how to automatically generate
bounding box annotations from the saliency maps and compare the generated
bounding boxes with previously published results. After this, we evaluate
several SOTA object detectors on the generated bounding box dataset.
Note that saliency map here refers to the mean attention map of our
proposed method, as also denoted in \secref{Saliency-Generation-Method}.

\subsection{Automatic Bounding Box Generation\label{subsec:Automatic-Bounding-Box}}

Since it is common in object detection tasks to infer bounding boxes,
a logical step is to generate bounding boxes for the dataset in order
to use SOTA object detection algorithms. For that, Saralajew et al.~\cite{saralajew2021pvdn-dataset}
already proposed an algorithm as well as a proper metric to measure
the quality of the automatically generated boxes given the keypoint
annotations. We show that by generating binary maps using our visual
saliency approach with the keypoints as fixation points, we can derive
higher-quality bounding boxes than the previously proposed approach.

We shortly recap the proposed bounding box metric by Saralajew et
al.~\cite{saralajew2021pvdn-dataset}. The metric phrases the problem
of detecting and classifying light instances of oncoming vehicles
at night as an object detection problem with binary classification,
meaning whether the object is a relevant light instance (direct or
indirect) or it is not. The primary goal of this bounding box metric
is to make predicted bounding boxes comparable with the ground-truth
keypoints. This is in contrast to the commonly performed comparison:
predicted bounding boxes are evaluated by comparing the predicted
with ground-truth bounding boxes using a measure based on the intersection-over-union
value~\cite{Liu2019}. In summary, the proposed metric follows the
idea that
\begin{itemize}
\item each bounding box should span around exactly one keypoint, and
\item each keypoint should lie within exactly one bounding box.
\end{itemize}
The following detection events are introduced:
\begin{itemize}
\item True positive: The light instance is described (at least one bounding
box spans over the keypoint);
\item False positive: The bounding box describes no light instance (no keypoint
lies within the bounding box);
\item False negative: The light instance is not described (no bounding box
spans over the keypoint).
\end{itemize}
With that, precision, recall, and F-score can be computed in the usual
manner. Additionally, in order to quantify the bounding box quality
of true-positive events, the following quantities are defined:
\begin{itemize}
\item $n_{K}(b)$: The number of keypoints in a bounding box $b$ that is
a true positive;
\item $n_{B}(k)$: The number of true-positive bounding boxes that cover
the same keypoint $k$.
\end{itemize}
With that, the quality measures can be computed as the averages over
all individual values by
\[
q_{K}=\frac{1}{N_{B}}\sum_{b}\frac{1}{n_{K}(b)}\in\left[0,1\right]
\]

and
\[
q_{B}=\frac{1}{N_{k}}\sum_{k}\frac{1}{n_{B}(k)}\in\left[0,1\right],
\]
where $N_{B}$ and $N_{K}$ are the total number of true-positive
bounding boxes and annotated keypoints, respectively. For example,
in an image with several keypoints, $q_{K}$ is low if a large bounding
box spans over the whole image since it will then capture several
keypoints. On the other hand, if many bounding boxes overlap and thus
span over the same keypoint, $q_{B}$ decreases. The overall quality
is determined by $q=q_{K}\cdot q_{B}\in\left[0,1\right]$.

Our bounding box generation pipeline consists of the following steps:
\begin{enumerate}
\item For each keypoint in an image, calculate the saliency map using the
keypoint as the fixation point (seed; see \secref{Saliency-Generation-Method}).
\item Use a blob detection approach to derive the bounding box coordinates
of each blob in the saliency map. Sometimes, this results in several
overlapping boxes.
\item For all possible combinations of bounding boxes derived from each
keypoint annotation, we calculate the F-score and the quality metric
$q$. Our first decision criterion is the F-score since we want to
capture as many keypoints as possible. Here, the recall can be used
equivalently, as precision is always 1.0, which is caused by the fact
that boxes generated with our saliency method cannot be false positive
since bounding boxes are only created if there is a keypoint. The
second decision criterion is the quality metric. Since out of the
combinations with the highest F-score, we want to retrieve the one
with the highest quality. As the metric is designed to penalize if
several boxes contain the same keypoint, choosing a combination with
a high quality value will lead to non-overlapping bounding boxes.
\end{enumerate}
\begin{figure*}
\begin{centering}
\includegraphics[width=0.8\paperwidth]{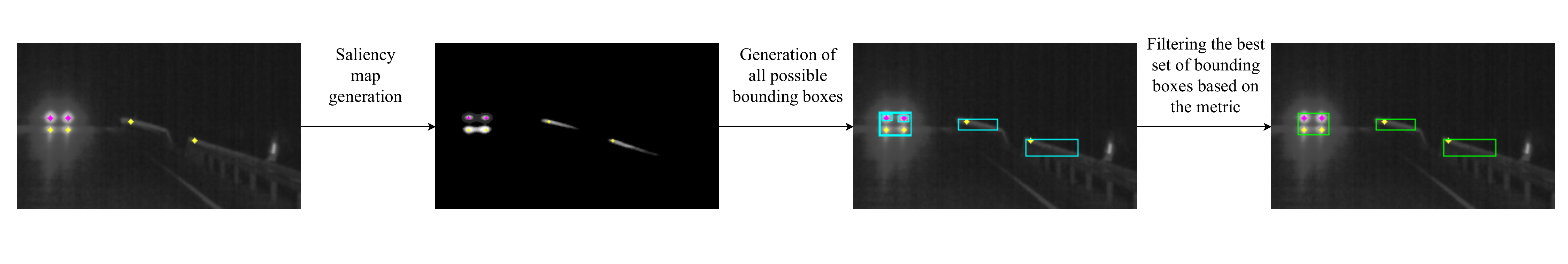}
\par\end{centering}
\caption{Bounding box generation pipeline using Boolean map saliency and the
ground-truth keypoints.\label{fig:bbox-pipeline} The first image
shows the original scene with the light instances annotated via keypoints
(pink: direct, yellow: indirect). In the first step, the saliency
map for each keypoint is generated. For better visualization, the
image shows the cumulated saliency map for all keypoints. Then, for
each saliency map, the bounding boxes are determined. Finally, the
combination of bounding boxes yielding the highest F-score and highest
quality score is selected.}
\end{figure*}
An example of these steps is shown in \figref{bbox-pipeline}. We
derive the optimal parameter setting for the saliency map generator
by optimizing the product of the F-score and the quality $q$ on the
whole PVDN dataset using a Tree-structured Parzen Estimator approach
as it was also done by Ewecker et al.~\cite{ewecker2021provident}.
This can be considered as a standard hyperparameter tuning\emph{.}
Thus, saliency maps for the whole dataset are generated \emph{using
the same parameter setting}.

\begin{table*}
\caption{Comparison of bounding box generation methods and performances of
trained object detectors (mean $\pm$ standard deviation).\label{tab:comparison-bbox-methods}}

\centering{}%
\begin{tabular}{ccccccc}
\toprule 
Model & Precision & Recall & F-score & $q$ & $q_{K}$ & $q_{B}$\tabularnewline
\midrule
Bounding Box Generation~\cite{saralajew2021pvdn-dataset} & 1.00 & 0.69 & 0.81 & 0.42 & 0.42$\thinspace\pm\thinspace$0.24 & 1.00$\thinspace\pm\thinspace$0.00\tabularnewline
YoloV5s~\cite{saralajew2021pvdn-dataset} & 0.99 & 0.66 & 0.80 & 0.37 & 0.38$\thinspace\pm\thinspace$0.20 & 0.98$\thinspace\pm\thinspace$0.09\tabularnewline
YoloV5x~\cite{saralajew2021pvdn-dataset} & 1.00 & 0.68 & 0.81 & 0.37 & 0.38$\thinspace\pm\thinspace$0.20 & 0.98$\thinspace\pm\thinspace$0.08\tabularnewline
\midrule
Bounding Box Generation~\cite{ewecker2021provident} & 1.00 & 0.87 & 0.93 & 0.70 & 0.70$\thinspace\pm\thinspace$0.30 & 1.00$\thinspace\pm\thinspace$0.00\tabularnewline
YoloV5s~\cite{ewecker2021provident} & 0.98 & 0.67 & 0.80 & 0.67 & 0.70$\thinspace\pm\thinspace$0.31 & 0.97$\thinspace\pm\thinspace$0.12\tabularnewline
YoloV5x~\cite{ewecker2021provident} & 0.99 & 0.76 & 0.86 & 0.67 & 0.69$\thinspace\pm\thinspace$0.30 & 0.98$\thinspace\pm\thinspace$0.10\tabularnewline
\midrule
Bounding Box Generation (ours) & 0.99 & 0.93 & 0.96 & 0.86 & 0.86$\thinspace\pm\thinspace$0.25 & 1.00$\thinspace\pm\thinspace$0.00\tabularnewline
YoloV5s (ours) & 0.96 & 0.65 & 0.78 & 0.87 & 0.89$\thinspace\pm\thinspace$0.23 & 0.98$\thinspace\pm\thinspace$0.10\tabularnewline
YoloV5x (ours) & 0.97 & 0.58 & 0.73 & 0.86 & 0.87$\thinspace\pm\thinspace$0.24 & 1.00$\thinspace\pm\thinspace$0.04\tabularnewline
\end{tabular}
\end{table*}
The results of the proposed saliency-based bounding box generation
approach are shown in \tabref{comparison-bbox-methods}. With our
method for generating bounding boxes based on the keypoint annotations,
we clearly outperform the previous approaches both in terms of F-score
and bounding box quality~\cite{ewecker2021provident,saralajew2021pvdn-dataset}.
With a bounding box quality score of 0.86, we see that the bounding
boxes, which are predicted correctly, are of very high quality. Specifically,
the improvement of $q_{K}$ shows that a single bounding box often
matches with exactly one keypoint, which indicates that small bounding
boxes separate keypoints that are located close to each other. A high
bounding box quality is desirable since it better reflects the original
human annotations, meaning that for each annotated keypoint there
exists in the optimal case exactly one unique derived bounding box. 

\subsection{SOTA Object Detector Evaluation}

We trained the SOTA object detection algorithm YoloV5\footnote{\url{https://github.com/ultralytics/yolov5}}
on the newly generated bounding boxes using the small (YoloV5s) and
large version (YoloV5x) to evaluate a run-time and detection performance
optimized algorithm, respectively. We used the Adam optimizer with
an initial learning rate of 0.01, weight decay of 0.0005, a batch
size of 16 and 32 for YoloV5x and YoloV5s, respectively, trained for
200 epochs each, and evaluated on the official PVDN test dataset.

Training YoloV5 on our saliency-based bounding boxes produces competitive
detection performance results compared to the benchmark by Saralajew
et al.~\cite{saralajew2021pvdn-dataset}. However, we achieve slightly
lower detection performances than Ewecker et al.~\cite{ewecker2021provident}.
This is most likely caused by the fact that our ground-truth bounding
boxes are often smaller and thus harder to detect than the ones created
with the approach by Ewecker et al.~\cite{ewecker2021provident}:
They achieve a significantly lower bounding box quality, indicating
that bounding boxes often span across several keypoints and, thus,
are bigger than ours.

In terms of bounding box quality, we surpass the previous methods
by Saralajew et al.~\cite{saralajew2021pvdn-dataset} and Ewecker
et al.~\cite{ewecker2021provident}. YoloV5, as our SOTA object detection
algorithm, is able to discriminate the different light instances much
better when trained on our saliency-based bounding box annotations.
By increasing the quality of the bounding boxes, the task inherently
becomes more difficult since a high bounding box quality indicates
that each keypoint has a single corresponding bounding box and only
a few bounding boxes span over multiple keypoints. Thus, the algorithm
has to distinguish between neighboring light instances and cannot
simply predict a larger bounding box spanning over several keypoint
annotations. This explanation provides a rationale for why YoloV5
performs slightly worse compared to Ewecker et al.~\cite{ewecker2021provident}
in terms of F-score, although being trained on better ground-truth
bounding boxes. For future autonomous systems, it might be crucial
to predict such high-quality bounding boxes, especially when several
light instances in an image have to be associated with its sources
(i.\,e., the vehicle that caused the light reflection). For that,
it is crucial to distinguish precisely between different unique light
instances and not mix them together as a single instance, which often
happens with the previously published methods.

Looking at the quality measures, it becomes obvious that $q_{B}$
converges to 1.0 in all cases. Considering the definition of $q_{B}$
and the bounding box generation methods, this behavior is natural.
The quality $q_{B}$ becomes 1.0 if each ground truth keypoint is
covered by not more than one bounding box, meaning that bounding boxes
do not overlap and multiple bounding boxes do not span over the same
keypoint. For the bounding box generation algorithm proposed by Saralajew
et al.~\cite{saralajew2021pvdn-dataset} and Ewecker et al.~\cite{ewecker2021provident}
overlapping bounding boxes are mitigated by the non-maximum suppression
at the end of the bounding box generation pipeline. In our saliency-based
method, saliency maps that would normally overlap are merged together
to a single box. When training YoloV5 variants on those bounding boxes,
they filter out overlapping candidates by the final non-maximum suppression
as well as learn to infer non-overlapping bounding boxes, therefore,
also resulting in $q_{B}$ values that converge to 1.0.

%% file: 06_discussion_conclusion.tex
\section{Conclusion and Future Work\label{sec:Conclusion}}

Following up on the previous work of Saralajew et al.~\cite{saralajew2021pvdn-dataset},
we presented an approach to generate various object representations
based on sparse keypoint annotations using visual saliency in order
to detect oncoming vehicles at night before they are actually directly
visible. With our method, the fuzzy and unclear borders of light reflections
can be properly modeled, thus, giving a more natural representation.
We showed that when using our context-aware saliency approach, the
task can easily be phrased as an object detection problem. The proposed
approach is able to automatically generate high-quality bounding boxes
based on the annotated keypoints describing the direct and indirect
light instances. On the PVDN dataset, we set a new benchmark for automatically
deriving bounding box annotations from human keypoint annotations.
We also show that, when trained on our generated bounding boxes, the
SOTA object detection algorithm YoloV5 is able to achieve superior
results compared to previous works. This is especially important when
building systems where it is necessary to find the correspondence
between light artifacts and their emitting source, for example, when
several vehicles are already visible, and it has to be determined
to which vehicle a light artifact belongs. In this case, having precise
and high-quality descriptions of the light artifacts is key.

In future work, we plan to use the saliency maps generated with our
approach to develop algorithms that can directly handle the fuzzy
boundaries of light reflections and that can deal with uncertain object
borders. For that, our proposed saliency method builds the foundation
by providing a framework to annotate objects with unclear boundaries.

In summary, in this paper, we provided further perspectives and tools
that rely on visual saliency to tackle the problem of provident vehicle
detection at night. At this point, we want to emphasize that although
the methods are evaluated on the specific automotive use case, we
are confident that they can be transferred to other domains where
fuzzy and objectively non-definable object boundaries are present,
such as biological image data of fluorescence microscopy.